\documentclass[11pt,letterpaper]{article}

\usepackage[margin=1.5in,top=1.1in,bottom=1.15in]{geometry}
\usepackage[T1]{fontenc}
\usepackage{newtxtext,newtxmath}
\usepackage{microtype}
\usepackage{graphicx}
\usepackage{booktabs}
\usepackage{amsmath}
\usepackage[hyphens]{url}
\usepackage[numbers,sort&compress]{natbib}
\usepackage{caption}
\usepackage[colorlinks=true,linkcolor=blue!55!black,citecolor=blue!55!black,urlcolor=blue!55!black]{hyperref}
\title{From Waste to Discovery: Recycling High-Scored Unexposed Candidates in Taobao Feed Recommendation}

\author{
    Ge Fan\textsuperscript{\rm 1}\thanks{Additional contact: \texttt{ge.fan@outlook.com}.},
    Kai Meng\textsuperscript{\rm 1},
    Huiping Chu\textsuperscript{\rm 1}\thanks{Corresponding author: \texttt{yueqi.chp@alibaba-inc.com}.},
    Yuning Jiang\textsuperscript{\rm 2},
    Han Zhu\textsuperscript{\rm 2},
    and Bo Zheng\textsuperscript{\rm 2}\\[6pt]
    {\small\textsuperscript{\rm 1}Taobao \& Tmall Group of Alibaba, Hangzhou, China}\\
    {\small\textsuperscript{\rm 2}Taobao \& Tmall Group of Alibaba, Beijing, China}\\
    {\small\{fange.fg, mengkai.meng, yueqi.chp, mengzhu.jyn, zhuhan.zh, bozheng\}@alibaba-inc.com}
}

\date{}

\begin{document}

\maketitle

\begin{abstract}
Helping users discover novel interests is essential for the long-term health of industrial recommender systems, yet the dominant remedy---boosting the exposure of under-explored (\emph{discovery}) content---often backfires in production. We report the deployment of URED (Unexposed Recycling for Efficient Discovery), a training-free recall-stage system that has been serving the Taobao feed recommendation system at Alibaba for over six months, covering tens of millions of daily active users. URED is built on an insight validated in our production data: naive exploration boosting fails because it injects cold, information-poor candidates that ranking models systematically over-estimate, yielding no significant discovery gain while putting engagement at risk. In contrast, items that were \emph{scored highly by the ranker but never exposed}---latent signals routinely discarded in industrial pipelines---constitute a candidate pool that is simultaneously high-relevance and reliably predictable. URED aggregates such candidates over a 30-day horizon and re-injects them at recall time through a lightweight pipeline spanning offline storage, distributed aggregation, and low-latency online serving, adding less than 1\% serving latency via a single $O(1)$ lookup per request with no model retraining. Large-scale online A/B experiments show that URED significantly improves discovered categories by +3.47\% and content richness by +8.47\% while preserving engagement: completion video views rise significantly by +2.62\%, and total video views move positive (+0.36\%, not statistically significant). Ablations confirm that both design pillars are necessary: shortening the aggregation window to one day collapses the discovery gain to a non-significant $+0.55\%$, and replacing model-score filtering with random discovery injection yields no significant benefit while trending negative on engagement ($-0.17\%$). Compared across production recall channels, URED delivers a 13.9$\times$ higher discovery rate than the platform average---6.3$\times$ the strongest nearline recall baseline---at engagement quality on par with the platform average. We share deployment details and lessons learned to guide practitioners facing the exploration--exploitation dilemma under strict resource constraints.
\end{abstract}

\section{Introduction}

Recommender systems---a flagship application of industrial AI---serve billions of requests per day through a cascaded pipeline of recall, ranking, and re-ranking optimized primarily for short-term engagement~\cite{chen2022meta, liu2024at4ctr, fan2026uniboost, zhang2026dream}. This exploitation-heavy design progressively narrows the content a user sees, producing the well-documented filter-bubble effect: recommendations become homogeneous, long-tail content is starved of exposure, and users lose opportunities to develop new interests~\cite{liu2019real, fan2022field, kim2025sequentially}. On our short-video platform, this manifests concretely: for a typical active user, the vast majority of daily impressions concentrate in a small set of familiar categories, while \emph{discovery categories}---those the user has not encountered in the recent past---account for only 6.2\% of exposed content platform-wide.

The standard industrial remedies fall into two families. The first family enriches models: multi-objective ranking, reinforcement-learning-based exploration, or LLM-augmented generation~\cite{lin2025order, jia2025learn, shao2026ugr}. These approaches raise training cost, inference latency, and infrastructure complexity---often prohibitively so in resource-constrained production environments. The second family bypasses models entirely: heuristic \emph{exploration boosting} that force-injects under-explored content into user feeds. This is cheap and simple, which explains its popularity; however, as we quantify in this paper, it also fails to deliver.

\textbf{Why naive boosting fails.} Our production analysis reveals a dual failure of boosted discovery content. Because boosted candidates are typically \emph{cold} with respect to the user---the model has little or no interaction signal connecting the user to the candidate's category---the ranking model falls back toward population priors, producing predictions that are simultaneously over-estimated relative to realized engagement and unreliable in ordering. The consequence is measurable: in our controlled online experiment, injecting randomly selected discovery candidates (including cold-start items) produced no significant discovery improvement ($+0.12\%$) and moved total video views negative ($-0.17\%$, not significant). Naive boosting buys nothing measurable and carries engagement risk.

\textbf{Our approach.} URED (Unexposed Recycling for Efficient Discovery) starts from a resource that industrial pipelines generate and discard at enormous scale: \emph{unexposed samples}, items that passed recall, were scored by the ranking model, but were never shown to the user due to position constraints and inter-candidate competition. Among discovery content, high-scored unexposed items occupy a privileged position. Under the working assumption that the ranker is (approximately) unbiased given sufficient input signal, a high score carries two pieces of information at once: the item is \emph{relevant} (the score estimates true engagement probability), and the prediction is \emph{trustworthy} (a model starved of signal regresses toward the prior and cannot produce a confidently high score). High-scored unexposed discovery items are therefore exactly the opposite of boosted cold items---they are the discovery candidates the model demonstrably understands.

URED operationalizes this insight as a lightweight, training-free recall channel. It ingests top-scored unexposed samples after each ranking pass, aggregates them per user over a 30-day sliding window using distributed batch infrastructure, filters them with a configurable discovery criterion (category unseen in the past $d$ days), and serves the resulting candidates from an in-memory store at recall time with sub-millisecond lookup. No model is retrained; no serving architecture is modified; measured latency overhead is below 1\%.

\textbf{Deployment and results.} URED has been deployed in production in the Taobao feed recommendation system for over six months, serving tens of millions of daily active users.
Online A/B experiments demonstrate that URED significantly improves discovered categories by +3.47\% and content richness by +8.47\% while engagement is preserved: completion views improve significantly by +2.62\%, and total views move positive (+0.36\%, not statistically significant)---to our knowledge a rare instance of a pure data-pipeline intervention improving discovery without sacrificing engagement. Ablation experiments validate each design pillar: a 1-day aggregation window retains only 16\% of the discovery gain (non-significant), and random injection without model-score filtering yields no significant improvement and trends negative on engagement. Channel-level analysis shows that 85.9\% of URED-exposed content belongs to discovery categories---13.9$\times$ the platform average and 6.3$\times$ the strongest nearline recall baseline---at engagement quality (valid-view rate) 1.03$\times$ the platform average.

\textbf{Contributions.} This paper makes three contributions aligned with the goals of deployed-application research:
(1) a production-validated diagnosis of why naive exploration boosting fails, tracing the failure to the ranking model's systematic over-estimation of cold discovery candidates---a model-calibration insight that motivates recycling the model's own high-confidence predictions;
(2) URED, a deployed, training-free recall augmentation system that recycles high-scored unexposed samples, with a data architecture spanning offline storage, distributed temporal aggregation, and low-latency serving;
(3) large-scale online evidence---including ablations and cross-channel comparisons---quantifying the necessity of each design choice, together with lessons learned from development, deployment, and operation.

\section{Why Naive Exploration Boosting Fails}
\label{sec:motivation}

We first formalize the failure mode that motivated URED, then present the design pivot that follows from it.

\subsection{Setting and Definitions}

Let $\hat{y}(u,i)$ denote the ranking model's predicted engagement probability for user $u$ and item $i$, and let $r(u,i)$ denote the true (realized) engagement probability. An item category is a \emph{discovery category} for $u$ if it has not been exposed to $u$ within the preceding $d$ days; items in such categories are \emph{discovery items}, denoted $\mathcal{D}$. After each ranking pass, items that were scored but not shown form the \emph{unexposed set}. We partition the discovery space into
\begin{align*}
\mathcal{D}^{\text{high}}_{\text{unexp}} &= \{ i \in \mathcal{D} : \hat{y}(u,i) \ge \tau,\ i \text{ never exposed to } u \},\\
\mathcal{D}_{\text{other}} &= \mathcal{D} \setminus \mathcal{D}^{\text{high}}_{\text{unexp}},
\end{align*}
where $\mathcal{D}_{\text{other}}$ contains cold-start discovery items (never scored for $u$) and low-scored discovery items.

\subsection{The Dual Failure of Cold Discovery Candidates}

Ranking models in production are trained overwhelmingly on exposed impressions. For a cold pair $(u,i)$---no interaction history linking $u$ to $i$'s category---the model's per-user signal is absent and its prediction regresses toward a population prior. This produces two coupled pathologies. First, \emph{over-estimation}: population priors reflect the average engaged user of a category, which is optimistic for a user who has never chosen to engage with it; the resulting predictions exceed realized engagement. Second, \emph{unreliable ordering}: with predictions clustered near the prior, the model cannot meaningfully rank cold candidates against each other, so any selection among them is effectively arbitrary.

Naive boosting inherits both pathologies wholesale: it selects candidates precisely from the cold region where the model over-promises and cannot discriminate. The user receives content that looks acceptable to the pipeline but underperforms in reality---engagement drops, and because the injected items are poorly chosen, discovery barely improves. Our Variant~B experiment (Section~\ref{sec:ablation}) confirms this outcome in a controlled online test, and the channel-level analysis (Section~\ref{sec:channel}) shows its item-level signature: discovery-oriented channels that bypass model confidence sit below platform-average engagement quality.

\textbf{Direct offline evidence.} This online failure has a measurable offline signature. Table~\ref{tab:calib} quantifies the two pathologies in production logs, using PCOC (predicted-over-actual clicks) as the calibration metric~\cite{deng2020calibrating} and RMSE as the ranking-error metric. Discovery candidates exhibit substantially larger over-estimation (PCOC $1.147$ vs.\ $1.042$, a $+10\%$ relative gap) and higher prediction error (RMSE $0.368$ vs.\ $0.325$, $+13\%$) than non-discovery candidates---exactly the coupled pathologies the mechanism above predicts. Naive boosting makes both worse on the same logging pipeline: applied to discovery candidates, PCOC and RMSE further rise by $+8.2\%$ and $+2.7\%$ relative, because boosting multiplies already-overestimated scores and admits higher-error candidates into the recall pool. The maximization-bias literature~\cite{fan2023calibration} formalizes this amplification: any positive prediction gap on a top-$K$ selected set grows monotonically once selection depends on the prediction itself. The ablation in Section~\ref{sec:ablation} (Variant~B) and the channel-level comparison (Section~\ref{sec:channel}, Cross-match at $0.77\times$ valid-PV rate) are the online behavioral signatures of this same offline calibration gap.

\begin{table}[t]
\centering
\caption{Ranker calibration and ranking error on production logs, stratified by candidate region. Discovery candidates are simultaneously over-estimated and harder to rank; naive boosting amplifies both pathologies---the coupled failure motivating URED.}
\label{tab:calib}
\small
\setlength{\tabcolsep}{4pt}
\begin{tabular}{lrr}
\toprule
Candidate region & PCOC & RMSE \\
\midrule
Non-discovery                   & $1.042$ & $0.325$ \\
Discovery                       & $1.147$ & $0.368$ \\
Discovery (naive boost)             & $1.241$ & $0.378$ \\
\bottomrule
\end{tabular}
\end{table}

\textbf{URED's design pivot.} The calibration gap in Table~\ref{tab:calib} points at its own remedy: rather than injecting \emph{any} discovery content---the region where PCOC is highest and RMSE is largest---the system should recycle the \emph{high-scored} subset of unexposed discovery candidates. Two properties make this region privileged. First, \emph{relevance}: a confidently high predicted score on an item the ranker demonstrably understands is an honest signal of true engagement probability, not the population-prior guess it falls back to on cold pairs. Second, \emph{predictability}: a confidently high score itself certifies that the model possessed sufficient per-user input signal, so its predictions there carry low error variance; high scores act as a built-in filter for information sufficiency. The two properties coincide only in the high-scored unexposed region---which industrial pipelines already compute on every request and throw away---and the empirical premise that they do coincide is what the experiments in Section~\ref{sec:experiments} will put to three tests. One further production constraint completes the design space: standard exposure deduplication prohibits re-showing previously exposed items, so the unexposed set is not merely preferable but the only admissible pool for re-injection.

\section{System Design}
\label{sec:system}

\begin{figure}[t]
\centering
\includegraphics[width=0.9\textwidth]{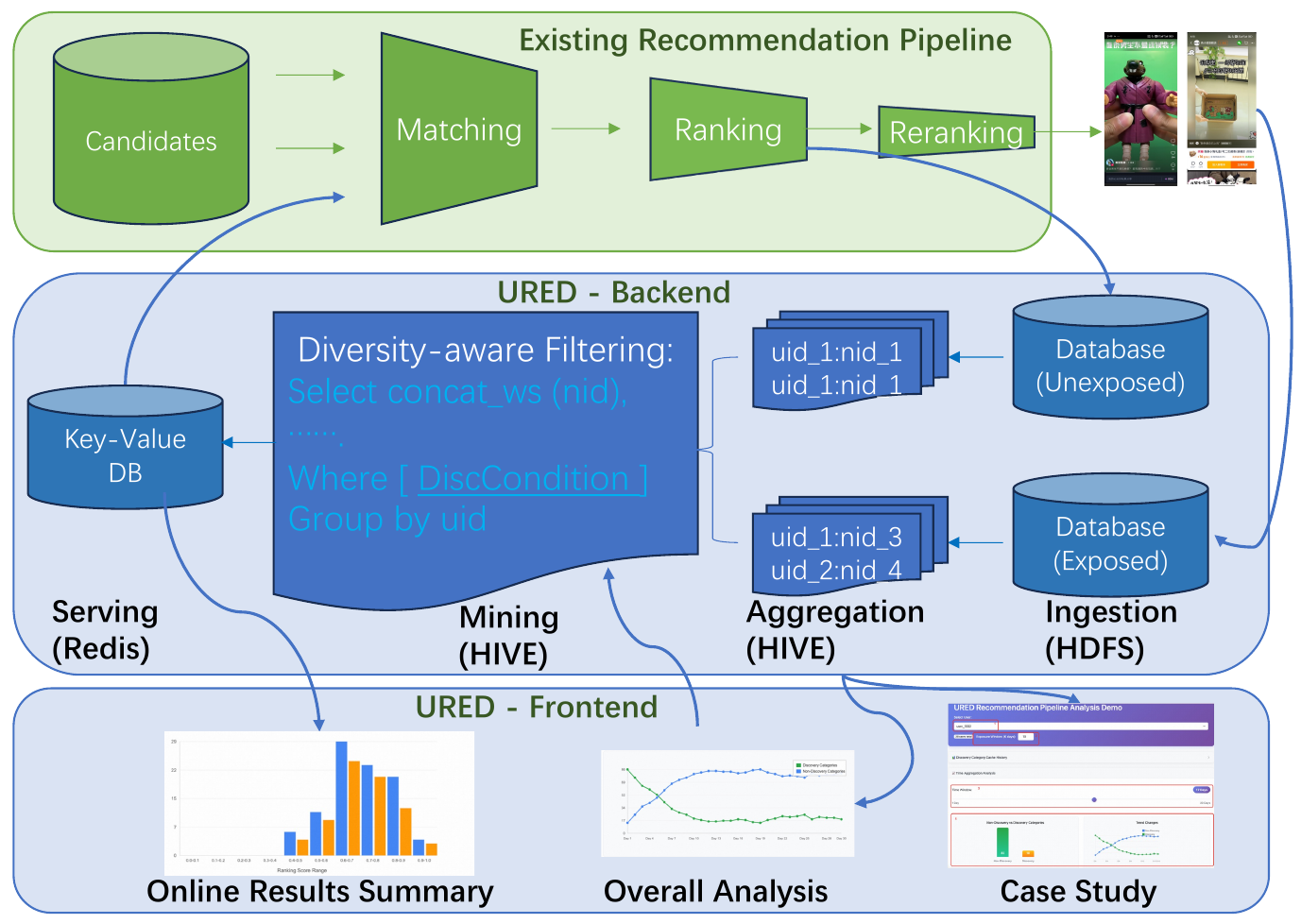}
\caption{Overview of URED. The green section is the existing recommendation pipeline; the blue section is URED, which recycles top-scored unexposed samples into an additional recall channel.}
\label{fig:framework}
\end{figure}

URED is an augmentation module attached to the recall stage of a standard three-stage industrial pipeline (Figure~\ref{fig:framework}). Its backend comprises four stages, each matched to the volume, velocity, and access pattern of the data it handles. We describe each stage together with the design decisions made during development.

\subsection{Stage 1: Sample Ingestion}
After each ranking pass, scored-but-unshown items are identified. Retaining every unexposed sample is infeasible---the ranker scores hundreds of candidates per request---so we retain the top-$k_1$ ($k_1 = 200$) highest-scoring candidates per session, enriched with category, timestamp, and component sub-scores, and persist them asynchronously to a distributed file store (e.g., HDFS) with minute-level latency. The top-$k_1$ truncation is not merely a storage optimization: the high-score region is precisely where the ranker's predictions are most trustworthy---Table~\ref{tab:calib} shows RMSE $0.325$ on non-discovery candidates versus $0.368$ on discovery ones---so truncation doubles as a quality filter. Exposed impressions are logged in parallel for downstream joins.

\subsection{Stage 2: Temporal Aggregation}
Daily batch jobs on a distributed warehouse engine (e.g., Hive/Spark) aggregate each user's unexposed history over a sliding window of $t$ days. Samples are deduplicated by item ID, ordered by recency, and truncated to the top-$k_2$ ($k_2 = 10{,}000$) per user. We set $t = 30$: our ablation (Section~\ref{sec:ablation}) shows that a 1-day window forfeits 84\% of the discovery gain, because too few high-scored unexposed discovery candidates accumulate within a single day to form a meaningful recall supplement. In parallel, the aggregation maintains each user's exposed-category set over the past $d$ days, which drives the discovery filter in the next stage.

\subsection{Stage 3: Discovery-Oriented Mining}
Joining the unexposed pool against the exposed-category set, URED retains candidates whose categories have not been exposed to the user in the past $d$ days ($d$ is configurable per product goal). This business-aligned novelty criterion directly targets the platform's discovery metric. An internal web dashboard lets operators tune $d$, inspect per-user exposure trajectories, and simulate the effect of alternative policies on candidate coverage before deployment; we found this human-in-the-loop step essential for aligning the filter with evolving campaign goals.

\subsection{Stage 4: Online Serving}
Filtered candidates are indexed by user ID in an online key-value store (e.g., Redis). At request time, URED is queried as one additional recall channel alongside embedding-based retrieval~\cite{fan2022pppne, liu2025llmemb}, item-to-item collaborative filtering~\cite{chen2019preference, zeng2021collaborative, wang2026pi2i}, and cross-domain matching~\cite{fan2022mv, ge2025behavior}; up to $k$ candidates are merged into the unified pool per request (we deploy $k=1000$; Section~\ref{sec:ablation} tunes $k$) and flow through unchanged ranking and re-ranking. Lookup is $O(1)$ per request, and production latency statistics accumulated over the long-horizon A/B experiments show the end-to-end serving overhead staying below 1\% on the treatment buckets relative to control. No model retraining or serving-stack modification is required, which was decisive for deployment approval.

\section{Deployment}
\label{sec:deployment}

URED is deployed in the Taobao feed recommendation system, a large-scale short-video platform, and has been serving production traffic for over six months. The platform serves tens of millions of daily active users and handles billions of requests per day, with peak traffic exceeding ten thousand QPS; URED participates in the recall stage of this hot path on every request.

URED's resource overhead splits into an online and an offline part. Online, serving the channel increases cluster CPU by only $+0.15\%$ and memory by $+0.53\%$: a few hundred CPU cores and a modest in-memory key-value footprint serve hundreds of millions of users, far below the compute and memory cost of the recall and ranking model services it augments. Offline, the daily batch pipeline involves only lightweight business logic---deduplication and merging---and finishes within tens of minutes at a parallelism of at most one thousand cores; because it is scheduled in the idle windows of the platform's existing batch infrastructure, it requires no additional compute capacity ($+0.0\%$ CPU and memory). Total storage grows by merely $+0.46\%$, and the persisted candidate pool doubles as a backup of online serving data, supporting fault recovery.
The system is operated by the existing recommendation infrastructure team without dedicated headcount: offline stages run as scheduled batch jobs in off-peak hours, and the serving path reuses the platform's standard key-value-store operations tooling. Candidate freshness, channel exposure share, and per-channel engagement quality are monitored daily; the discovery threshold $d$ and injection quota are the two operator-facing knobs, adjusted through the dashboard described in Section~\ref{sec:system}.

\section{Online Experiments}
\label{sec:experiments}

We evaluate URED through large-scale online A/B experiments on production traffic, where tens of millions of daily active users generate billions of requests per day (peak QPS above ten thousand). Users were randomly assigned to a control bucket (baseline pipeline, 10\% of traffic) and treatment buckets (baseline + URED recall channel, 5--10\% of traffic), with traffic allocated by the platform's global coordination framework~\cite{fan2026uniboost}. Every reported experiment ran for more than one week under this allocation; statistical significance is assessed with a $Z$-test~\cite{student1908probable} enhanced by CUPED variance reduction~\cite{deng2013improving} and annotated per table.
We report engagement metrics---total video views (VV) and completion VV---and discovery metrics: discovered categories (count of newly encountered categories), average discovered categories per user, richness (distinct content diversity of exposure), and average richness per user; the discovery criterion uses $d=15$ days throughout. Owing to company data policy, absolute values of engagement-quality metrics are normalized by the platform-wide average where noted. The experiments are designed to put the design pivot of Section~\ref{sec:motivation} to three tests: (i) URED-injected items should sustain engagement quality, since a pool selected from a miscalibrated region would drag it below the platform average; (ii) removing the score filter must destroy the lift, since random injection cannot substitute for low-error predictions; and (iii) completion-view gains should materialize, since genuine relevance is what makes users watch content through rather than skip it.

\subsection{Overall A/B Results}

\begin{table}[t]
\centering
\caption{Relative changes from deploying URED (treatment vs.\ control). All metrics pass the significance test except Total VV, whose positive movement does not reach significance.}
\label{tab:overall}
\small
\setlength{\tabcolsep}{4pt}
\begin{tabular}{lr@{\hspace{1em}}lr}
\toprule
Metric & Change & Metric & Change\\
\midrule
Total VV & $+0.36\%$ & Completion VV & $+2.62\%$\\
Disc.\ Categories & $+3.47\%$ & Avg.\ Disc.\ Categories & $+1.89\%$\\
Richness & $+8.47\%$ & Avg.\ Richness & $+1.69\%$\\
\bottomrule
\end{tabular}
\end{table}

Table~\ref{tab:overall} summarizes the main experiment. URED improves every discovery metric---most notably +8.47\% richness and +3.47\% discovered categories---all of which pass the significance test. Engagement is preserved and improves where it matters most: completion VV rises significantly by +2.62\%, and total VV moves positive by +0.36\%, though the latter does not reach significance. These results validate the \emph{relevance} premise (test~i): the valid-PV rate at $1.03\times$ platform average (Section~\ref{sec:channel}) confirms that URED-injected items sustain engagement quality. They also validate the \emph{predictability} premise (test~iii): the significant completion-view gain (+2.62\%) shows that users watch URED content through rather than skim, consistent with low-error predictions.

\subsection{Ablation: Both Design Pillars Are Necessary}
\label{sec:ablation}

URED rests on two design choices: (i) \emph{long-horizon aggregation} ($t=30$ days) and (ii) \emph{model-score filtering} (recycling only high-scored unexposed candidates). We isolate each with controlled online variants. Variant~A keeps model-score filtering but shrinks the window to one day. Variant~B keeps the injection quota but selects discovery candidates at random (including cold-start items), removing the model-score filter---this variant doubles as a faithful implementation of naive exploration boosting. Re-exposing previously shown items is architecturally prohibited by exposure deduplication, so no ``exposed'' variant exists; URED operates in the only admissible complement space.

\begin{table}[t]
\centering
\caption{Ablation of URED's two design pillars (relative to the no-URED control). Only URED-Full's discovery gain is statistically significant ($\dagger$); all other movements, including every engagement change, do not reach significance (n.s.).}
\label{tab:ablation}
\small
\setlength{\tabcolsep}{3pt}
\begin{tabular}{llrr}
\toprule
Variant & Candidate source / window & Total VV & Disc.\ Cat.\\
\midrule
URED-Full & unexposed high-score, 30d & $+0.36\%$ & $+3.47\%^{\dagger}$\\
Variant A & unexposed high-score, 1d & $+0.02\%$ & $+0.55\%$\\
Variant B & random disc.\ (incl.\ cold) & $-0.17\%$ & $+0.12\%$\\
\bottomrule
\end{tabular}
\end{table}

Table~\ref{tab:ablation} yields three findings. \textbf{(1) Long-horizon aggregation is necessary.} Shrinking the window from 30 days to 1 day collapses the discovery gain from +3.47\% (significant) to +0.55\% (n.s.)---only 16\% of the full effect---with engagement movement vanishing as well. A single day simply does not accumulate enough high-scored unexposed discovery candidates per user to be useful. \textbf{(2) Model-score filtering is necessary.} Random discovery injection (Variant~B) fails to improve discovery significantly (+0.12\%, n.s.) and moves total VV negative ($-0.17\%$, n.s.). This is the dual failure of Section~\ref{sec:motivation} surfacing in production: cold candidates that the model over-estimates and cannot rank displace better content, dragging engagement down without buying discovery. This finding directly validates the \emph{predictability} premise: the score filter is what enforces low prediction-error variance, and discarding it destroys both relevance and predictability at once---any-discovery injection is not a substitute for high-confidence selection. \textbf{(3) Only the full design delivers.} URED-Full is the only variant with a statistically significant benefit---the discovery gain---and a positive engagement movement. Variant~B exemplifies why exploration boosting has historically been unpopular with product teams. The contrast shows that the engagement--discovery trade-off is not inherent to exploration; it is an artifact of injecting candidates the model does not understand. Its root cause is the offline calibration gap quantified in Table~\ref{tab:calib}: Variant~B injects from the region where PCOC is highest ($1.147\to 1.241$ under boosting) and RMSE is largest ($0.368\to 0.378$), so the online failure directly reflects this offline gap.

\begin{figure}[t]
\centering
\includegraphics[width=0.62\textwidth]{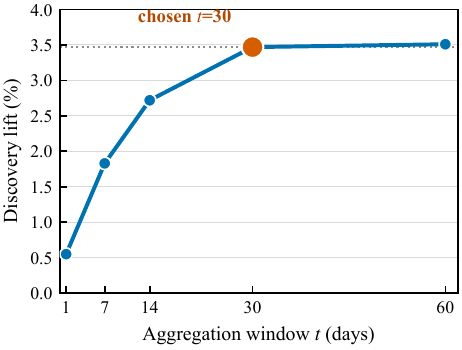}
\caption{Discovery lift as a function of the aggregation window $t$. The gain rises steeply then saturates, with a knee at $t=30$; extending to 60 days adds negligible benefit while increasing storage.}
\label{fig:window}
\end{figure}

To characterize the aggregation horizon more finely, we sweep $t \in \{1, 7, 14, 30, 60\}$ days (Figure~\ref{fig:window}); all sensitivity configurations here and below run as concurrent parallel buckets over the same period, ruling out temporal confounds. The discovery gain rises steeply from +0.55\% ($t=1$) to +2.72\% ($t=14$), then saturates: +3.47\% at $t=30$ and only +3.51\% at $t=60$. The knee at $t=30$ shows that roughly a month of history suffices to accumulate the high-scored unexposed discovery candidates URED can act on; longer windows add stale candidates and storage without meaningful gain. We therefore fix $t=30$ as the operating point balancing benefit and cost.

\begin{figure}[t]
\centering
\includegraphics[width=0.62\textwidth]{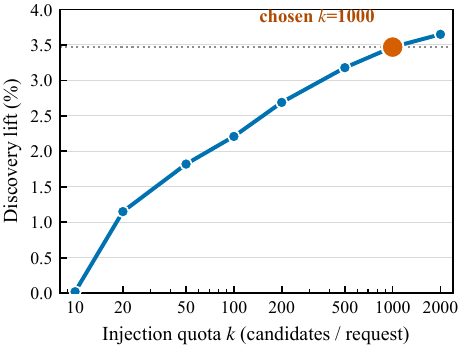}
\caption{Discovery lift as a function of the per-request injection quota $k$ (log axis). Gains grow with strongly diminishing returns; the deployed point $k=1000$ captures most of the attainable lift.}
\label{fig:topk}
\end{figure}

We likewise sweep the per-request injection quota $k$---the number of URED candidates merged into the recall pool---over $k \in \{10,\dots,2000\}$ (Figure~\ref{fig:topk}). The lift grows with strongly diminishing returns: +0.02\% at $k=10$, +2.69\% at $k=200$, +3.47\% at $k=1000$, and only +3.65\% at $k=2000$. The deployed point $k=1000$ captures most of the attainable gain; beyond it, extra injected candidates are increasingly marginal and begin to crowd the recall pool, so we fix $k=1000$.

\textbf{Supply-side evidence for the deployed $(t,k)=(30,1000)$.} The sensitivity curves above show \emph{what} happens as $t$ and $k$ vary; supply-side statistics explain \emph{why}. URED's pool is a finite resource: each user generates only a limited supply of high-scored unexposed discovery candidates per day, and the two hyperparameters jointly determine how much of it the pipeline harvests.

\begin{figure}[t]
\centering
\begin{minipage}[b]{0.48\textwidth}
\centering
\includegraphics[width=\linewidth]{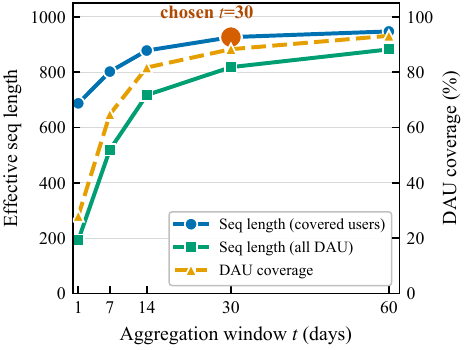}
\end{minipage}
\hfill
\begin{minipage}[b]{0.48\textwidth}
\centering
\includegraphics[width=\linewidth]{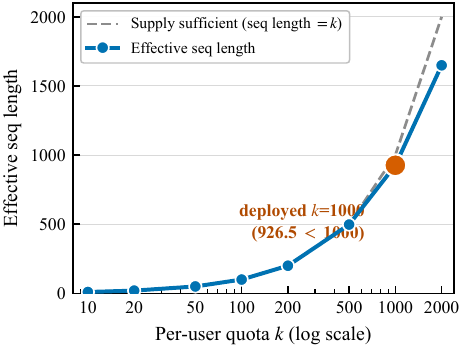}
\end{minipage}
\caption{Supply-side evidence for the deployed hyperparameters. (a)~Window $t$ versus DAU coverage (fraction of daily active users with a non-empty pool), per-covered-user sequence length, and DAU-average (real) sequence length; coverage saturates around $t=30$. (b)~Per-user quota $k$ versus effective sequence length; the dashed reference line marks the supply-sufficient regime (seq length $=k$), from which the curve detaches at $k=1000$.}
\label{fig:supply}
\end{figure}

Figure~\ref{fig:supply}(a) reports DAU coverage (the fraction of daily active users whose pool becomes non-empty), the effective sequence length averaged over covered users, and the DAU-average (real) sequence length. Coverage rises steeply from 28.0\% ($t=1$) to 88.3\% ($t=30$) and then flattens to 93.1\% ($t=60$): extending the window mostly recruits occasionally active users rather than deepening existing pools. From 30 to 60 days the per-covered-user length grows only $+2.2\%$ (926.5$\to$947.2), doubling storage and batch computation for $+0.04$~pp online; $t=30$ is the coverage--cost elbow. Figure~\ref{fig:supply}(b) compares the effective sequence length with the supply-sufficient reference (seq length $=k$). Up to $k=500$ the curve hugs the reference---virtually every covered user possesses at least $k$ high-scored candidates---but detaches at $k=1000$ (average 926.5$<$1000), reaching only 1649.4 at $k=2000$: doubling the quota grows pools by merely 78\% because the high-scored unexposed supply is nearly exhausted within the 30-day window. Moreover, marginal candidates beyond rank 1000 carry lower model scores and hence weaker confidence in the ranker's predictions, consistent with the $+0.18$~pp online delta from 1000 to 2000 (Figure~\ref{fig:topk}). In short, $(t,k)=(30,1000)$ sits at the joint elbow: 88\% of DAU are already covered, per-user supply is close to saturated at ${\approx}926$ candidates, and further enlarging either hyperparameter doubles its cost for less than 0.2~pp of online gain.

\subsection{Channel-Level Quality Analysis}
\label{sec:channel}

To situate URED among production recall channels, we compare per-channel exposure quality on live traffic. The baselines cover the platform's main recall portfolio: \emph{Embedding-Based}, two-tower retrieval trained with sampling-bias correction~\cite{yi2019sampling}; \emph{I2I}, item-to-item collaborative filtering over a large-scale product graph~\cite{yang2020large}; \emph{Cross-match}, rule-based cross-domain matching built on personalized item-based retrieval~\cite{wang2026pi2i}; and \emph{MNR}, a multi-scenario nearline framework that retrieves from users' real-time behaviors~\cite{ma2024simple}. \emph{Valid PV rate} is the probability that an exposed item is watched beyond a fixed duration threshold (engagement quality); \emph{discovery rate} is the fraction of a channel's exposed items belonging to discovery categories. Valid PV rates are normalized by the platform-wide average (OVERALL $=1.00$) per data policy; discovery rates are reported in absolute terms.

\begin{table}[t]
\centering
\caption{Recall-channel comparison on production traffic. Valid PV rate is normalized by the platform average; discovery rate is absolute. MNR is the nearline-recall baseline.}
\label{tab:channel}
\small
\setlength{\tabcolsep}{5pt}
\begin{tabular}{lcc}
\toprule
Channel & Valid PV rate (norm.) & Discovery rate\\
\midrule
Embedding-Based & 1.19 & 7.4\%\\
I2I & 1.13 & 9.4\%\\
Cross-match & 0.77 & 6.8\%\\
MNR & 1.13 & 13.6\%\\
Platform overall & 1.00 & 6.2\%\\
\textbf{URED} & \textbf{1.03} & \textbf{85.9\%}\\
\bottomrule
\end{tabular}
\end{table}

\begin{figure}[t]
\centering
\includegraphics[width=0.62\textwidth]{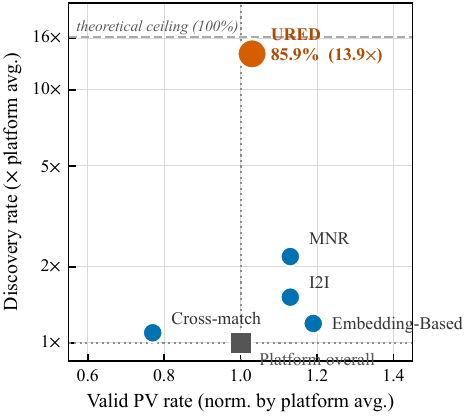}
\caption{Quality--discovery operating points of production recall channels.}
\label{fig:qd}
\end{figure}

Table~\ref{tab:channel} and Figure~\ref{fig:qd} deliver the paper's central quantitative message. URED's exposed content is 85.9\% discovery---13.9$\times$ the platform average and 6.3$\times$ MNR (13.6\%), the strongest nearline recall baseline---while its valid PV rate (1.03) remains at the platform average. MNR achieves a higher per-item valid PV rate (1.13) but at a fraction of URED's discovery concentration; URED's modest per-item quality gap relative to MNR is the price of a 6$\times$ higher discovery concentration, and crucially URED never falls below platform-average quality. The comparison against Cross-match (0.77) further shows that discovery-oriented mining by the model's own high scores dominates rule-based cross-domain matching on both dimensions. It also closes the loop with Section~\ref{sec:motivation}: URED's candidates deliver the engagement the ranker promised---the empirical counterpart of the relevance and predictability premises underlying URED's design---whereas rule-based candidates that bypass model confidence (Cross-match) underdeliver on it, mirroring the offline PCOC/RMSE gap reported in Table~\ref{tab:calib}.

Two complementary views of the discovery lift are worth highlighting. In absolute terms, over 85\% of what URED injects is discovery content; the theoretical ceiling is 100\%, since every candidate in URED's pool is filtered by the discovery criterion offline, and the $\approx$14~pp gap is attributable to the latency between offline filtering and online key-value serving rather than to the mechanism itself. In relative terms, URED achieves 13.9$\times$ the platform average while holding efficiency metrics unchanged---conventional channels cannot separate discovery from engagement cost this way.

Taken together, the channel-level evidence completes the validation of all three tests previewed at the start of this section. (i) Engagement quality preserved at $1.03\times$ platform average while 85.9\% of URED exposure is discovery content confirms the \emph{relevance} premise: the ranker's high scores on these candidates translate into realized engagement, so relevance is genuine rather than an artifact of over-estimation. (ii) Variant~B's random injection---which bypasses the score filter---produces no significant discovery gain and trends negative on engagement (Table~\ref{tab:ablation}), confirming the \emph{predictability} premise: it is the low prediction-error variance of high-scored candidates, not merely their presence in the discovery set, that makes URED work. (iii) The significant completion-view gain (+2.62\%) connects both premises to one behavioral result: users watch URED content through because it is both relevant and reliably ranked---something no other discovery channel, rule-based or model-bypassing, achieves in our production data.

\subsection{Who Benefits: User-Segment Analysis}
\label{sec:segment}

\begin{figure}[t]
\centering
\includegraphics[width=0.62\textwidth]{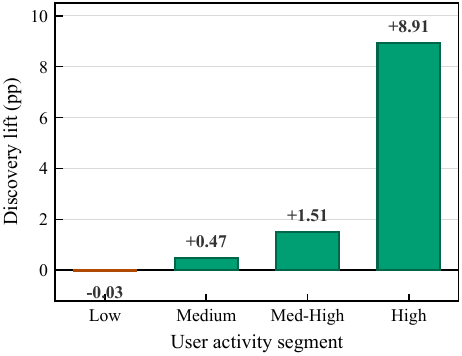}
\caption{Discovery lift in percentage points (experiment $-$ baseline, on normalized proportions) by user activity segment. Gains grow monotonically with activity; low-activity users are essentially unchanged.}
\label{fig:segment}
\end{figure}

To understand who benefits, we decompose the (normalized) discovery proportion by user activity segment and report the experiment$-$baseline difference in percentage points (pp; Figure~\ref{fig:segment}). The lift grows monotonically with activity: $-0.03$~pp (low, negligible), $+0.47$~pp (medium), $+1.51$~pp (med-high), and $+8.91$~pp (high). This pattern is exactly what URED's mechanism predicts. Since URED's pool is built purely from unexposed samples, its size scales with how much ranking traffic a user generates: medium- and high-activity users accumulate a rich supply of high-scored-but-unexposed candidates over the 30-day window, so their recycled pools are dense and yield large discovery gains. Low-activity users, by contrast, have sparse interaction behavior and thus a thin pool, so the near-zero lift is an expected consequence of the data source rather than a failure of the mechanism. More importantly, for low-activity users discovery of new categories is not the primary objective in the first place---retention and core-viewing satisfaction take priority---so URED deliberately defers this population to cold-start channels (Section~\ref{sec:lessons}). In short, URED concentrates its benefit where the supply of high-confidence unexposed candidates is abundant and where discovery is actually wanted, without harming low-activity users.

\section{Lessons Learned and Discussion}
\label{sec:lessons}

\textbf{Measure the failure before building the fix.} Our first instinct, like much of the industry, was to boost discovery exposure directly. Quantifying the over-estimation of cold discovery candidates---and watching Variant~B return no significant gain with a negative engagement trend in a controlled test---redirected the effort from ``inject more'' to ``select better,'' which ultimately defined URED. We recommend that teams facing filter-bubble pressure first audit the calibration of their ranker on cold content; the audit is cheap and its result dictates the correct intervention.

\textbf{The model's own scores are an underused exploration asset.} The high-score region of unexposed samples is computed on every request by every industrial ranker, yet is almost universally discarded. Recycling it required no new model and no new signal---only data-lifecycle engineering. The cost profile follows directly: a single $O(1)$ key-value lookup per request keeps end-to-end latency overhead below 1\%, and the offline cost is scheduled batch aggregation reusing existing data-warehouse capacity (Section~\ref{sec:deployment}). Against naive boosting---which is equally training-free---URED is strictly dominant at identical cost, delivering significant discovery gain where boosting delivers none. The same ``waste-to-value'' pattern generalizes beyond category discovery to \emph{freshness} (re-injecting high-scored items never shown), \emph{creator or supplier fairness} (surfacing under-represented creators), and \emph{long-tail exposure}; the single requirement is a relevance signal trustworthy for never-exposed items (Table~\ref{tab:calib}). URED inherits the ranker's blind spots: its candidate quality is bounded by the model's ability to score unseen items, so segments where the ranker is miscalibrated receive less benefit and the system defers to other channels there. Cold-start users are the sharpest case---their thin interaction history produces a thin unexposed pool---which is why we blend URED with cold-start recall in operation~\cite{zhang2022quickskill}.

\textbf{Patient accumulation beats per-request cleverness.} Per-request, high-scored unexposed discovery candidates are scarce; over a month they are abundant. Exploration pipelines should be designed around \emph{accumulation horizons}, not just \emph{injection quotas}: the value of the channel comes from gathering a trustworthy candidate pool over time, and practitioners should expect short-window replications to yield substantially smaller gains.

\textbf{Screen hyperparameters with offline supply indicators, not online sweeps.} Hyperparameters such as the window $t$ and quota $k$ control how much candidate supply the pipeline harvests, so their effect is largely visible \emph{before} any online experiment. Two cheap offline indicators proved highly predictive: UV coverage (the fraction of DAU whose pool becomes non-empty) and effective sequence-length growth (how fast per-user pools deepen). Whenever coverage has plateaued or the pool length saturates against its supply ceiling, further expansion buys little online. Screening configurations offline with these indicators narrowed a combinatorial space to one or two candidates and saved most of a full online sweep's A/B budget.

\textbf{Operator-facing controllability accelerates deployment.} Improving a production recommender involves not only engineers but also product managers and operations staff. A black-box pipeline makes cross-team alignment difficult---when outcomes disappoint, no one can trace why. URED's dashboard addresses this by exposing the discovery threshold $d$, per-user trajectories, and simulated policy impacts. Technical staff use these views to tune hyperparameters; non-technical partners use the same interface to verify system behavior and to flag category over-concentration. This transparency converted the launch discussion from a leap of faith into a parameter negotiation.

\section{Related Work}

\textbf{Discovery and diversity in recommendation.} Prior work combats homogenization via multi-objective ranking~\cite{li2019multi,peng2024reconciling}, multi-criteria preference modeling~\cite{fan2021predicting,fan2023improving}, topic- and social-aware user-interest modeling~\cite{chen2023topic, chen2023neural}, representation-enrichment architectures~\cite{wu2024enhancing}, and LLM-based generation~\cite{wei2024llmgr,lin2025order}; mechanism-design analyses formalize the ecosystem-level diversity--exploitation balance~\cite{boutilier2024recommender}, and causal-deconfounding methods promote diversity via counterfactual exposure~\cite{zhang2026diversity}. A complementary line treats discovery as explicit exploration: bandit- and RL-based methods allocate impressions to uncertain items at an explicit engagement cost~\cite{ashkan2015optimal}, while deployed systems re-matchmake already-evaluated candidates to improve satisfaction~\cite{fan2024cupid, wang2026champ} or inject Bayesian posteriors to absorb cold start and non-stationarity in large-scale search~\cite{bayescns2025}. These raise training/serving cost or pay an exploration price; URED is complementary---it explores only where the ranker is already confident, purely at the data layer with no model changes.

\textbf{Utilizing unexposed samples in recommender systems.} Industrial rankers score far more candidates than they expose. Nearline retrieval, pioneered by Glare~\cite{glare2013} and deployed at scale in multi-scenario frameworks such as MNR~\cite{ma2024simple}, re-retrieves on recent user behavior; CroPS likewise harvests cross-perspective positive samples---query reformulation and non-click engagement---to train dense retrieval in short-video search~\cite{crops2026}. Knowledge distillation compresses ranking models for short-video recommendation~\cite{mikd2025}, and full-flow datasets such as RecFlow release unexposed samples for offline study~\cite{liu2025recflow}. Counterfactual and exposure-bias correction also consume non-exposure, but only to debias training~\cite{wu2019pd}; conversely, deployed systems re-surface \emph{previously-exposed} content for retention~\cite{jiang2026save}. All of these use unexposed samples as a \emph{training-time} signal; URED instead promotes high-scored unexposed candidates directly to \emph{serving-time} recall, requiring no retraining, counterfactual modeling, or extra inference.

\section{Conclusion}

We presented URED, a deployed, training-free system that turns a universally discarded byproduct of industrial ranking---high-scored unexposed samples---into a production recall channel for discovery. Operating at tens-of-millions-DAU scale with a single $O(1)$ key-value lookup per request and no retraining, URED rests on the premise that high-scored unexposed candidates are simultaneously relevant and reliably predictable. Online experiments---showing +3.47\% discovered categories and +8.47\% richness alongside preserved engagement---confirm that the engagement--discovery trade-off attributed to exploration is instead an artifact of injecting content the model cannot vouch for. Ablations quantify the necessity of long-horizon aggregation and model-score filtering, and channel-level analysis shows URED reaching a quality--discovery operating point (1.03$\times$ average engagement quality at 85.9\% discovery concentration) unmatched by conventional channels. We hope the system design and lessons reported here help practitioners pursue exploration without sacrificing the metrics their platforms live by.

\bibliographystyle{unsrtnat}
\bibliography{short}

\end{document}